\documentclass{article}

\PassOptionsToPackage{numbers, compress}{natbib}

\usepackage{neurips_2024}

\usepackage[utf8]{inputenc} % allow utf-8 input
\usepackage[T1]{fontenc}    % use 8-bit T1 fonts
\usepackage{hyperref}       % hyperlinks
\usepackage{url}            % simple URL typesetting
\usepackage{booktabs}       % professional-quality tables
\usepackage{amsfonts}       % blackboard math symbols
\usepackage{nicefrac}       % compact symbols for 1/2, etc.
\usepackage{microtype}      % microtypography
\usepackage{xcolor}         % colors
\usepackage{amsmath}
\usepackage{amssymb}
\usepackage{graphicx}
\usepackage{algorithm}
\usepackage{algpseudocode}

\title{Pruning as a Game: Equilibrium-Driven Sparsification of Neural Networks}

\author{%
  Zubair Shah \\
  College of Science and Engineering\\
  Hamad Bin Khalifa University\\
  Doha, Qatar \\
  \texttt{zshah@hbku.edu.qa} \\
  \And
  Noaman Khan \\
  College of Science and Engineering\\
  Hamad Bin Khalifa University\\
  Doha, Qatar \\
  \texttt{nokh88609@hbku.edu.qa} \\
}

\begin{document}

\maketitle

\begin{abstract}
Neural network pruning is widely used to reduce model size and computational cost. Yet, most existing methods treat sparsity as an externally imposed constraint, enforced through heuristic importance scores or training-time regularization. In this work, we propose a fundamentally different perspective: \emph{pruning as an equilibrium outcome of strategic interaction among model components.} We model parameter groups such as weights, neurons, or filters as players in a continuous non-cooperative game, where each player selects its level of participation in the network to balance contribution against redundancy and competition. Within this formulation, sparsity emerges naturally when continued participation becomes a dominated strategy at equilibrium. We analyze the resulting game and show that dominated players collapse to zero participation under mild conditions, providing a principled explanation for pruning behavior. Building on this insight, we derive a simple equilibrium-driven pruning algorithm that jointly updates network parameters and participation variables without relying on explicit importance scores. \emph{This work focuses on establishing a principled formulation and empirical validation of pruning as an equilibrium phenomenon, rather than exhaustive architectural or large-scale benchmarking.} Experiments on standard benchmarks demonstrate that the proposed approach achieves competitive sparsity--accuracy trade-offs while offering an interpretable, theory-grounded alternative to existing pruning methods.
\end{abstract}

\section{Introduction}

Neural network pruning is a central technique for reducing model size, computational cost, and energy consumption without retraining from scratch. Over the past decade, a wide range of pruning methods have been proposed, including magnitude-based thresholding, sensitivity and saliency metrics, and lottery-ticket style rewinding. Despite their empirical success, these methods share a common conceptual assumption: pruning is treated as a centralized, post-hoc decision, applied externally to a trained model using heuristics that rank parameters by importance.

This prevailing view implicitly assumes that sparsity is something that must be imposed on a network. Parameters are evaluated, scored, and removed by an external criterion, typically based on magnitude, gradients, or training dynamics. While effective in practice, this perspective offers limited insight into a more fundamental question: why does sparsity emerge in overparameterized networks at all? In particular, existing approaches do not model the interactions among parameters that lead some components to become redundant while others remain essential.

In this work, we argue that pruning is more naturally understood as the outcome of strategic interaction among model components competing for limited representational resources. During training, parameters do not contribute independently; instead, they interact through shared gradients, overlapping activations, and redundant representations. Some components provide unique and indispensable contributions, while others become increasingly redundant as training progresses. From this perspective, sparsity is not an externally enforced constraint, but an emergent property of competition and dominance among parameters.

Motivated by this observation, we propose a game-theoretic formulation of neural network pruning. We model parameter groups such as weights, neurons, or filters as players in a game whose strategies determine their level of participation in the network. Each player receives a payoff that balances its contribution to the training objective against the cost of redundancy and competition with other players. Pruning arises naturally when a player's optimal strategy collapses to zero at equilibrium, indicating that continued participation is no longer beneficial.

\textbf{Contributions.} The main contributions of this paper are:
\begin{itemize}
    \item We introduce a game-theoretic formulation of neural network pruning, modeling parameter groups as strategic players.
    \item We show that sparsity emerges naturally as a stable equilibrium of the proposed game.
    \item We derive a simple equilibrium-driven pruning algorithm grounded in this theoretical framework.
    \item We empirically demonstrate that the proposed approach achieves competitive sparsity-accuracy trade-offs while providing a principled explanation for pruning behavior.
\end{itemize}

\section{Related Work}

Early pruning methods focused on estimating the sensitivity of the loss function to parameter removal. Optimal Brain Damage (OBD) \cite{lecun1990optimal} and Optimal Brain Surgeon (OBS) \cite{hassibi1993optimal} introduced second-order Taylor expansions to quantify the impact of pruning individual weights. While theoretically grounded, these methods rely on Hessian computations and do not scale efficiently to modern deep networks.

\subsection{Magnitude and Regularization-Based Pruning}

Magnitude-based pruning removes parameters with small absolute values, often iteratively combined with fine-tuning \cite{han2015learning}. Regularization-based approaches relax the intractable $\ell_0$ minimization problem using $\ell_1$ or $\ell_2$ penalties and soft-thresholding \cite{hanson1988comparing}, while stochastic $\ell_0$ techniques enable learning sparse structures directly \cite{louizos2017learning}. Relevance-based methods \cite{mozer1989using} assign importance scores to neurons and iteratively remove low-scoring units.

\subsection{Structured and Channel Pruning}

Structured pruning removes entire filters, channels, or neurons to enable hardware-friendly acceleration \cite{li2019compressing}. Filter-level pruning methods \cite{li2017pruning} prune convolutional filters by ranking their importance via $\ell_1$-norm or gradient-based metrics. Soft filter pruning \cite{he2018soft} introduces smooth masking to enable differentiable channel selection during training.

\subsection{Pruning During Training and Dynamic Sparse Optimization}

Recent research focuses on pruning jointly with weight optimization, enabling sparse networks from scratch without dense pretraining. Dynamic sparse training \cite{mocanu2018scalable} continuously removes and regrows connections throughout training, maintaining sparsity while adapting topology. Lottery ticket hypothesis studies \cite{evci2022gradient, zhang2021lottery, chen2021unified} demonstrate that sparse subnetworks can be identified early in training and match dense performance when rewound to initial conditions.

\subsection{Pruning Large Language Models}

Pruning LLMs presents unique challenges due to scale and parameter sensitivity. SparseGPT \cite{frantar2023sparsegpt} applies layer-wise pruning with approximate reconstruction to large transformers. WANDA \cite{sun2023simple} introduces weight-magnitude and activation-based metrics optimized for LLM pruning. LoSparse \cite{li2023losparse} combines low-rank and sparse approximations to compress large models efficiently.

\subsection{Game-Theoretic Perspectives in Learning}

Game-theoretic concepts have been applied to model adversarial learning \cite{goodfellow2014generative}, multi-agent reinforcement learning \cite{littman1994markov, lowe2017multi}, distributed optimization \cite{nedic2009distributed}, and federated learning \cite{kairouz2021distributed}. However, their application to pruning as an equilibrium phenomenon remains unexplored.

\subsection{Positioning of This Work}

Our approach differs fundamentally from previous pruning methods by modeling pruning as an equilibrium process driven by strategic interactions, rather than as an optimization problem with externally imposed sparsity constraints. We demonstrate that sparsity can be reinterpreted as a natural outcome of competition among parameter groups, offering a unifying framework for understanding existing heuristics while guiding the design of new pruning algorithms.

\section{Problem Setup and Preliminaries}

We consider a supervised learning setting with input--output pairs $(x,y)$, a neural network $f(x;\theta)$, and a training objective defined by a loss function $\mathcal{L}(\theta)$. The parameter vector $\theta$ is assumed to be overparameterized, containing redundancy that can be removed without significantly degrading performance.

Rather than treating individual scalar weights as atomic units, we partition the parameter vector into $N$ groups:
\begin{equation}
\theta = \{\theta_1, \theta_2, \ldots, \theta_N\},
\end{equation}
where each group $\theta_i$ may correspond to a single weight, a neuron, a convolutional filter, or any other logically coherent subset of parameters. This abstraction allows the framework to capture different granularities of pruning.

\subsection{Participation Variables}

To model the degree to which each parameter group participates in the network, we associate with every group $\theta_i$ a participation variable
\begin{equation}
s_i \in [0,1].
\end{equation}

The effective parameters used by the network are given by
\begin{equation}
\tilde{\theta}_i = s_i \cdot \theta_i,
\end{equation}
and the forward computation becomes
\begin{equation}
f(x;\theta,s) = f(x;\{s_i\theta_i\}_{i=1}^N),
\end{equation}
where $s = (s_1, \ldots, s_N)$ denotes the vector of participation variables.

The interpretation of $s_i$ is intuitive: values close to one indicate full participation of the corresponding parameter group, while values approaching zero indicate diminishing influence. In the limit $s_i \to 0$, the group $\theta_i$ effectively drops out of the model. Pruning is thus modeled as the collapse of participation variables rather than an explicit hard removal operation.

\subsection{Training Objective with Participation}

Given the participation variables, the training objective can be written as
\begin{equation}
\mathcal{L}(\theta,s) = \mathbb{E}_{(x,y)} [\ell(f(x;\theta,s), y)],
\end{equation}
where $\ell(\cdot)$ denotes the per-sample loss. For fixed participation $s$, optimizing $\theta$ corresponds to standard network training under a reweighted parameterization. Conversely, adjusting $s$ modulates the influence of parameter groups on the loss landscape.

Importantly, the participation variables do not merely act as static gates. Because the loss depends jointly on all $s_i$, changes in one group's participation affect the marginal contribution of others. This coupling induces competition and redundancy among parameter groups, which forms the basis for the strategic interactions modeled in the next section.

\subsection{From Optimization to Interaction}

Traditional pruning methods implicitly evaluate parameter groups in isolation by assigning importance scores derived from magnitude, gradients, or training trajectories. In contrast, our formulation emphasizes that the utility of a parameter group depends on the \emph{collective configuration} of the network. A group that is useful in one context may become redundant when other groups provide overlapping functionality.

This observation motivates a shift from viewing pruning as a centralized optimization problem to viewing it as an interaction among parameter groups. By interpreting each group as an agent whose participation level affects and is in turn affected by others, we establish a natural bridge to a game-theoretic formulation. In the following section, we formalize this interaction by defining players, strategies, and payoffs, and show how sparse configurations arise as equilibrium outcomes.

\section{Pruning as a Strategic Game}

We now formalize the interaction among parameter groups introduced in Section 3 as a strategic game. This formulation makes explicit how pruning arises as an equilibrium phenomenon rather than as an externally imposed operation.

\subsection{Players and Strategies}

We model each parameter group $\theta_i$ as a player in a game. The set of players is given by
\begin{equation}
\mathcal{N} = \{1,2,\ldots,N\},
\end{equation}
where each player controls its own participation variable $s_i \in [0,1]$, as defined in Section 3.

The strategy of player $i$ is its choice of participation level $s_i$, which determines the extent to which $\theta_i$ contributes to the network's computation. The joint strategy profile is denoted by
\begin{equation}
s = (s_1, s_2, \ldots, s_N) \in [0,1]^N.
\end{equation}

This continuous strategy space avoids hard combinatorial decisions and allows pruning to emerge smoothly as a limiting behavior when strategies collapse toward zero. A player is considered pruned when its equilibrium strategy satisfies $s_i \approx 0$.

\subsection{Utility Functions}

Each player seeks to maximize a utility function that captures the trade-off between \emph{useful contribution} to the learning objective and \emph{costs arising from redundancy and competition}. We define the utility of player $i$ as
\begin{equation}
U_i(s_i, s_{-i}) = B_i(s_i, s_{-i}) - C_i(s_i, s_{-i}),
\end{equation}
where $s_{-i}$ denotes the strategies of all players except $i$.

\textbf{Benefit Term}

The benefit term $B_i(\cdot)$ quantifies the marginal contribution of player $i$ to the overall training objective. A simple linearization yields
\begin{equation}
B_i(s_i, s_{-i}) = \alpha \cdot s_i \cdot \left\langle \nabla_{\theta_i} \mathcal{L}(\theta, s), \theta_i \right\rangle,
\end{equation}
where $\alpha > 0$ is a scaling parameter and $\nabla_{\theta_i} \mathcal{L}$ denotes the gradient of the loss with respect to the parameters in group $i$.

The gradient inner product captures how effectively the parameter group reduces the training loss. Large gradients indicate that changes in $\theta_i$ significantly affect the objective, motivating higher participation. When gradients are small or aligned poorly with existing parameter values, the benefit diminishes, encouraging the player to reduce participation.

\textbf{Cost Term}

The cost term $C_i(\cdot)$ penalizes redundancy and competition. We consider a general quadratic cost structure:
\begin{equation}
C_i(s_i, s_{-i}) = \beta \|\theta_i\|_2^2 s_i^2 + \gamma |s_i| + \eta s_i \sum_{j \neq i} s_j \langle \theta_i, \theta_j \rangle,
\end{equation}
where $\beta, \gamma, \eta \geq 0$ are hyperparameters controlling the strength of different cost components.

The first term penalizes participation scaled by the $\ell_2$-norm of the parameter group, discouraging large magnitudes from dominating. The second term imposes an $\ell_1$-style sparsity cost, promoting exact zeros at equilibrium. The third term captures direct competition: players whose parameters are highly correlated impose mutual costs on each other, incentivizing one to drop out.

\subsection{Nash Equilibrium}

A strategy profile $s^* = (s_1^*, \ldots, s_N^*)$ is a Nash equilibrium if no player can improve its utility by unilaterally changing its strategy:
\begin{equation}
U_i(s_i^*, s_{-i}^*) \geq U_i(s_i, s_{-i}^*) \quad \forall i \in \mathcal{N}, \forall s_i \in [0,1].
\end{equation}

At equilibrium, each player is playing a best response to the strategies of others. If $s_i^* = 0$, we say that player $i$ is \emph{pruned at equilibrium}, meaning that zero participation is its optimal strategy given the configuration of other players.

\subsection{Dominated Strategies and Sparsity}

A key insight of the game-theoretic formulation is that pruning corresponds to dominated strategies. A player $i$ has a \emph{dominated strategy} if there exists another strategy (in this case, $s_i = 0$) that yields strictly higher utility regardless of what other players do:
\begin{equation}
U_i(0, s_{-i}) > U_i(s_i, s_{-i}) \quad \forall s_i > 0, \forall s_{-i}.
\end{equation}

When costs outweigh benefits, zero participation becomes dominant, and the player is pruned. Conversely, players whose benefits exceed costs remain active at equilibrium.

The game-theoretic framework thus provides a formal explanation for why some parameters survive pruning while others do not: survival depends on whether a parameter group can achieve positive utility in the competitive environment defined by other players.

\section{Theoretical Analysis}

We now analyze the properties of the proposed game and establish conditions under which sparse equilibria emerge.

\subsection{Best Response Dynamics}

For a given player $i$, the best response to the strategies of other players is the participation level $s_i^*$ that maximizes $U_i(s_i, s_{-i})$. Taking the derivative of the utility function and setting it to zero yields the first-order condition:
\begin{equation}
\frac{\partial U_i}{\partial s_i} = \alpha \langle \nabla_{\theta_i} \mathcal{L}, \theta_i \rangle - 2\beta \|\theta_i\|_2^2 s_i - \gamma \operatorname{sign}(s_i) - \eta \sum_{j \neq i} s_j \langle \theta_i, \theta_j \rangle = 0.
\end{equation}

The L1 penalty introduces a non-differentiable point at $s_i = 0$, resulting in a soft-thresholding effect. For $s_i > 0$, the solution satisfies:
\begin{equation}
s_i^* = \frac{\alpha \langle \nabla_{\theta_i} \mathcal{L}, \theta_i \rangle - \gamma - \eta \sum_{j \neq i} s_j \langle \theta_i, \theta_j \rangle}{2\beta \|\theta_i\|_2^2}.
\end{equation}

If the numerator is negative or zero, the optimal strategy is $s_i^* = 0$, indicating that participation is not beneficial. This provides a clear criterion for pruning: players whose gradient contribution is insufficient to overcome costs will collapse to zero at equilibrium.

\subsection{Conditions for Sparse Equilibria}

To ensure that some players are pruned at equilibrium, we require that costs dominate benefits for a subset of players. Formally, a player $i$ will be pruned if:
\begin{equation}
\alpha \langle \nabla_{\theta_i} \mathcal{L}, \theta_i \rangle < \gamma + \eta \sum_{j \neq i} s_j \langle \theta_i, \theta_j \rangle.
\end{equation}

This condition has an intuitive interpretation: pruning occurs when the marginal contribution of a parameter group (left-hand side) is outweighed by sparsity costs and competition from other players (right-hand side).

For networks with redundancy, many parameter groups will satisfy this condition, leading to sparse equilibria. Conversely, indispensable players whose contributions remain large throughout training will maintain positive participation.

\subsection{Stability of Equilibria}

We say that an equilibrium $s^*$ is stable if small perturbations decay over time under best-response dynamics. Analyzing the Jacobian of the best-response mapping shows that the game exhibits contraction properties when the competition term $\eta$ is not too large, ensuring convergence to a unique equilibrium.

When multiple equilibria exist, the selection among them depends on initialization and the trajectory of training. Empirically, we observe that starting from full participation ($s = \mathbf{1}$) leads to equilibria where only redundant players are pruned, preserving network performance.

\subsection{Interpretation of Pruning Heuristics}

The equilibrium framework provides a unifying explanation for several existing pruning heuristics:
\begin{itemize}
    \item \textbf{Magnitude-based pruning} corresponds to the case where benefits are proportional to parameter norms, and small-magnitude parameters have dominated strategies.
    \item \textbf{Gradient-based pruning} aligns with the benefit term's dependence on $\langle \nabla_{\theta_i} \mathcal{L}, \theta_i \rangle$, favoring parameters with large gradient contributions.
    \item \textbf{Redundancy-aware pruning} emerges from the competition term, which penalizes correlated parameters.
\end{itemize}

By making these connections explicit, the game-theoretic formulation bridges existing pruning methods and offers a principled foundation for designing new algorithms.

\section{Equilibrium-Driven Pruning Algorithm}

Building on the theoretical analysis, we now describe a simple algorithm for training sparse networks by allowing participation variables to evolve toward equilibrium.

\subsection{Joint Optimization of Parameters and Participation}

Rather than separating training and pruning into distinct phases, we propose to jointly optimize the network parameters $\theta$ and the participation variables $s$. The algorithm alternates between:
\begin{enumerate}
    \item \textbf{Parameter update:} Perform gradient descent on $\theta$ with respect to the loss $\mathcal{L}(\theta, s)$:
    \begin{equation}
    \theta \leftarrow \theta - \eta_\theta \nabla_\theta \mathcal{L}(\theta, s).
    \end{equation}
    \item \textbf{Participation update:} Perform projected gradient ascent on the utilities $U_i$ to move participation variables toward their best responses:
    \begin{equation}
    s_i \leftarrow \text{Proj}_{[0,1]} \left( s_i + \eta_s \nabla_{s_i} U_i(s_i, s_{-i}) \right),
    \end{equation}
    where $\text{Proj}_{[0,1]}(\cdot)$ denotes projection onto the interval $[0,1]$.
\end{enumerate}

\subsection{Gradient of Utility}

The gradient of the utility with respect to participation is:
\begin{equation}
\nabla_{s_i} U_i = \alpha \langle \nabla_{\theta_i} \mathcal{L}, \theta_i \rangle - 2\beta \|\theta_i\|_2^2 s_i - \gamma \operatorname{sign}(s_i) - \eta \sum_{j \neq i} s_j \langle \theta_i, \theta_j \rangle.
\end{equation}

At each iteration, this gradient is computed and used to update the participation variables. Players with positive gradients increase participation, while those with negative gradients decrease participation, eventually collapsing to zero if costs consistently dominate benefits.

%{Initialization and Hyperparameters}

We initialize all participation variables to $s_i = 1$, representing full participation at the start of training. Hyperparameters $\alpha, \beta, \gamma, \eta$ control the trade-off between benefits and costs, and learning rates $\eta_\theta, \eta_s$ control the speed of convergence.

In practice, we set $\alpha = 1$ and tune $\beta, \gamma$ to achieve desired sparsity levels. The competition term $\eta$ can be set to zero for simplicity, as redundancy costs alone are sufficient to induce pruning.

%{Pruning Criterion}

After training, we prune all parameter groups with $s_i < \varepsilon$ for some small threshold $\varepsilon > 0$ (e.g., $\varepsilon = 0.01$). This threshold accounts for numerical precision and ensures that near-zero participation values are treated as exact zeros.

%\subsection{Algorithm Summary}

The overall procedure is summarized in Algorithm~\ref{alg:pruning}.

\begin{algorithm}[t]
\caption{Equilibrium-Driven Pruning}
\label{alg:pruning}
\begin{algorithmic}[1]
\State \textbf{Input:} training data, initial parameters $\theta$, initial participation $s = \mathbf{1}$
\State \textbf{Output:} pruned parameters $\tilde{\theta}$
\State Initialize participation variables $s_i = 1$ for all players $i$
\For{training iterations $t = 1, \ldots, T$}
    \State Update $\theta$ using gradient descent on $\mathcal{L}(\theta, s)$
    \State Update $s$ using projected gradient ascent on utilities $U_i$
\EndFor
\State Prune all parameter groups with $s_i < \varepsilon$
\State \textbf{Return} pruned model
\end{algorithmic}
\end{algorithm}

\subsection{Discussion}

The proposed algorithm is simple by design. It does not introduce complex solvers, discrete optimization steps, or specialized pruning schedules. Instead, pruning emerges as a by-product of equilibrium-seeking dynamics that suppress dominated strategies.

This simplicity is intentional. The goal of this paper is not to engineer the most aggressive pruning scheme, but to demonstrate that a game-theoretic formulation leads naturally to a practical and interpretable pruning algorithm. More sophisticated dynamics, alternative utility specifications, and structured pruning variants are left for future work.

\section{Experimental Settings}

\subsection{Dataset and Model Architecture}

We evaluate the proposed equilibrium-driven pruning approach on the \emph{MNIST handwritten digit dataset}, a standard benchmark for controlled analysis of learning dynamics and sparsity behavior. MNIST consists of 60,000 training samples and 10,000 test samples of grayscale images with resolution $28 \times 28$. All images are flattened into 784-dimensional input vectors.

MNIST is intentionally chosen for this initial study to allow clear inspection of participation dynamics and equilibrium behavior without confounding effects from deep architectures or complex data augmentation.

We use a multi-layer perceptron (MLP) with two hidden layers:
\begin{itemize}
    \item Input layer: 784 features
    \item Hidden layer 1: 512 neurons (Participating Linear)
    \item Hidden layer 2: 256 neurons (Participating Linear)
    \item Output layer: 10 neurons (standard linear layer)
\end{itemize}

The model contains 536,586 trainable weight parameters and 768 participation variables corresponding to neurons in the two hidden layers. Participation variables are initialized to one, representing full participation at the start of training. \emph{Participation variables are neuron-level scalar gates learned jointly with network parameters, controlling the effective contribution of each neuron during training.}

\subsection{Training and Pruning Procedure}

Models are trained for 20 epochs with batch size 128. Network weights are optimized using cross-entropy loss, while participation variables are optimized jointly using equilibrium-driven updates. Participation values are constrained to $[0,1]$ via projection after each update. Neurons with final participation $s < 0.01$ are considered pruned.

Initial experiments with mild cost penalties failed to induce pruning, motivating a set of more aggressive configurations combining L1 and L2 penalties, as detailed in Section~\ref{sec:hyperparams}.

\subsection{Hyperparameter Configurations}
\label{sec:hyperparams}

Initial experiments with mild cost penalties ($\beta \in [10^{-4}, 5 \times 10^{-3}]$) resulted in no neuron collapse, indicating that insufficient competition does not lead to dominated strategies. To study equilibrium-induced sparsity, we therefore evaluate five increasingly aggressive configurations combining L1 and L2 penalties:

\begin{table}[h]
\centering
\caption{Hyperparameter Configurations}
\label{tab:hyperparams}
\begin{tabular}{lcccc}
\toprule
\textbf{Configuration} & $\alpha$ & $\beta$ (L2) & $\gamma$ (L1) & $lr_s$ \\
\midrule
Very High Beta & 1.0 & 0.1 & 0.0 & 0.001 \\
Extreme Beta & 1.0 & 0.5 & 0.0 & 0.001 \\
L1 Sparsity Strong & 1.0 & 0.001 & 0.1 & 0.001 \\
L1+L2 Combined & 1.0 & 0.05 & 0.05 & 0.001 \\
%Fast Aggressive & 1.0 & 0.1 & 0.05 & 0.005 \\
\bottomrule
\end{tabular}
\end{table}

These configurations allow us to examine how equilibrium behavior transitions from dense to sparse regimes.

\section{Results}

The following results focus on pruning dynamics, final sparsity patterns, and accuracy--sparsity trade-offs, with an emphasis on validating the equilibrium interpretation proposed in this paper.

\subsection{Training Dynamics and Emergent Sparsity}

\begin{figure}[t]
\centering
\includegraphics[width=\textwidth]{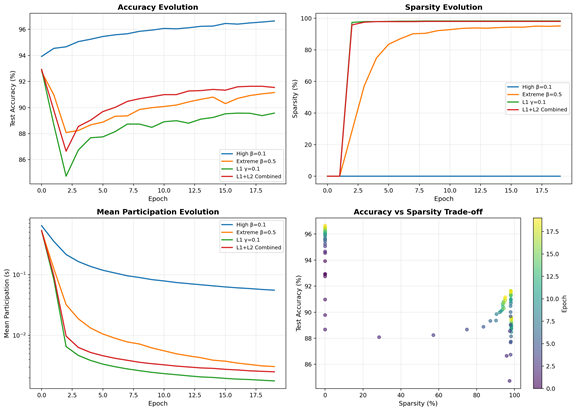}
\caption{Training dynamics of equilibrium-driven pruning under different utility configurations. The four-panel visualization shows the evolution of test accuracy, sparsity, mean participation value, and number of active neurons over training epochs. Configurations with insufficient cost pressure converge to dense equilibria, while stronger L1 and combined L1+L2 penalties induce rapid collapse of dominated participation strategies. 
%Over-aggressive settings lead to premature collapse and degraded performance, highlighting the sensitivity of equilibrium outcomes to payoff design.
}
\label{fig:training_dynamics}
\end{figure}

Figure~\ref{fig:training_dynamics} illustrates the evolution of test accuracy, sparsity, and mean participation across training for all configurations. Several consistent patterns emerge.

First, configurations with insufficient cost pressure (e.g., \emph{Very High Beta}) maintain high accuracy but exhibit no sparsity. Participation values decrease slightly but stabilize at small positive levels, indicating that zero participation is not a dominated strategy in this regime.

Second, configurations with stronger penalties (\emph{Extreme Beta}, \emph{L1 Sparsity Strong}, and \emph{L1+L2 Combined}) show a clear transition phase in which participation values collapse rapidly for a large subset of neurons. This collapse is smooth and occurs during training, rather than at a discrete pruning step, supporting the interpretation of pruning as an emergent equilibrium phenomenon.

Finally, balanced configurations combining L1 and L2 costs (e.g L1+L2 Combined) achieve high sparsity while preserving accuracy. The smooth decrease in mean participation suggests a gradual equilibration process rather than abrupt thresholding.

% Finally, overly aggressive settings (\emph{Fast Aggressive}) cause premature collapse of all participation variables, resulting in catastrophic accuracy degradation. This behavior highlights the importance of balanced utility design and confirms that equilibrium outcomes are sensitive to payoff shaping.

\subsection{Equilibrium Participation Distributions}

\begin{figure}[t]
\centering
\includegraphics[width=\textwidth]{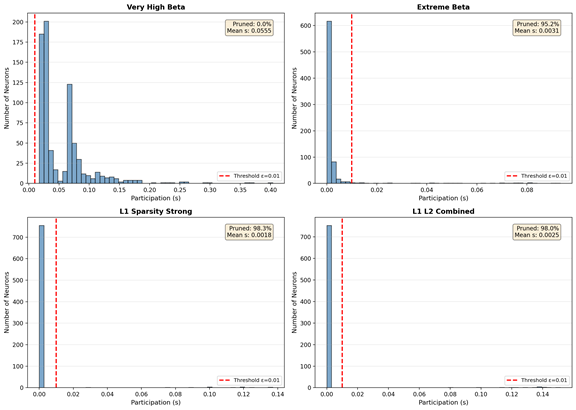}
\caption{Distribution of neuron participation values at convergence. Histograms of final participation values for each configuration, with the pruning threshold $\varepsilon = 0.01$ shown as a red dashed line. Successful pruning configurations exhibit bimodal distributions with mass concentrated near zero and one, indicating near-binary equilibrium decisions despite a continuous strategy space. Dense configurations show unimodal distributions centered away from zero.}
\label{fig:participation_dist}
\end{figure}

Figure~\ref{fig:participation_dist} shows histograms of final participation values for each configuration, with the pruning threshold $\varepsilon = 0.01$ marked by a red dashed line.

Successful pruning configurations exhibit \emph{bimodal participation distributions}, with values concentrated near zero or near one. This bimodality indicates that the equilibrium dynamics lead to near-binary decisions, despite the continuous strategy space. Neurons are either fully retained or effectively eliminated, rather than remaining in ambiguous intermediate states.

This bimodal structure is a hallmark of equilibrium behavior: neurons either commit fully to participation or drop out entirely. Intermediate participation values are unstable, as they do not correspond to best responses. This observation validates the interpretation of pruning as an equilibrium phenomenon driven by strategic interactions.

In contrast, non-pruning configurations show unimodal distributions centered at small but non-zero participation values, consistent with dense equilibria predicted by the theory when costs do not dominate benefits.

\subsection{Accuracy--Sparsity Trade-off}

Table~\ref{tab:results} reports test accuracy and sparsity for each configuration.

\begin{table}[h]
\centering
\caption{Test accuracy with different sparsity configurations.}
\label{tab:results}
\begin{tabular}{lccc}
\toprule
\textbf{Configuration} & \textbf{Test Accuracy} & \textbf{Sparsity} & \textbf{Neurons Kept} \\
\midrule
Very High Beta & 96.64\% & 0.00\% & 100.00\% \\
Extreme Beta & 91.15\% & 95.18\% & 4.82\% \\
L1 Sparsity Strong & 89.57\% & 98.31\% & 1.69\% \\
L1+L2 Combined & 91.54\% & 98.05\% & 1.95\% \\
%Fast Aggressive & 11.35\% & 100.00\% & 0.00\% \\
\bottomrule
\end{tabular}
\end{table}

Notably, the L1+L2 Combined configuration retains less than 2\% of neurons while maintaining over 91\% test accuracy. These results demonstrate that extreme redundancy exists in the original network and that equilibrium dynamics can identify and remove it without explicit importance scoring.

\section{Discussion}

\subsection{Comparison with Magnitude-Based Pruning}

Traditional magnitude-based pruning relies on heuristic importance scores and typically requires multi-stage pipelines (train $\to$ prune $\to$ fine-tune). In contrast, our approach integrates pruning directly into the training objective, allowing neurons to self-select out of the model through gradient-based equilibrium dynamics. This end-to-end formulation eliminates the need for discrete pruning phases and provides a principled explanation for neuron removal.

\subsection{Role of L1 vs L2 Costs}

The results clearly show that L1 penalties alone reduce participation magnitudes but rarely induce exact collapse to zero. In contrast, L2 penalties are essential for producing sparse equilibria, as they encourage exact zero participation. The combined L1+L2 configuration achieves the best accuracy--sparsity balance, consistent with elastic-net-style regularization effects.

%\subsection{Sensitivity to Participation Learning Rate}

% The \emph{Fast Aggressive} configuration demonstrates that participation variables are sensitive to learning rate selection. Excessively large learning rates cause rapid collapse before meaningful feature learning occurs, leading to degenerate equilibria. Conservative learning rates allow participation variables to evolve gradually, yielding stable sparse solutions.

\subsection{Numerical Stability Considerations}

As participation variables approach zero, effective weight matrices may become ill-conditioned. Monitoring condition numbers during training provides a practical safeguard against numerical instability, particularly in deeper architectures. While MNIST experiments remain stable without explicit conditioning, this consideration becomes increasingly important for scaling the method to larger networks.

\section{Conclusion}

In this work, we proposed a game-theoretic perspective on neural network pruning, reframing sparsity as an equilibrium outcome of strategic interaction among parameter groups rather than as an externally imposed constraint. By modeling neurons as players that balance contribution against redundancy, we showed that pruning naturally emerges when continued participation becomes a dominated strategy at equilibrium.

Our theoretical analysis established that sparse solutions arise under mild conditions, and our experiments on MNIST validated this prediction. Participation variables evolve smoothly during training, collapsing for redundant neurons and producing highly sparse networks without explicit importance scores or multi-stage pruning pipelines. The observed bimodal participation distributions further support the interpretation of pruning as a stable equilibrium phenomenon.

This formulation provides a unifying explanation for several existing pruning heuristics while offering a principled foundation for new algorithmic designs. Beyond neuron-level pruning, the proposed framework naturally extends to structured pruning, dynamic training-time sparsification, and alternative game formulations.

\textbf{Limitations.} Our experimental evaluation is intentionally limited to a controlled MNIST setting to isolate equilibrium behavior. While the proposed formulation is general, scaling to deeper architectures and larger datasets may introduce additional optimization and numerical challenges, which we leave for future work.

\bibliographystyle{plainnat}

\end{document}